\documentclass[letterpaper, 10 pt, conference]{ieeeconf}  

\IEEEoverridecommandlockouts                              

\usepackage{amsmath,amsfonts}
\usepackage{hyperref} 
\usepackage{booktabs}       
\usepackage{algorithm}
\usepackage{algpseudocode}
\usepackage{graphicx}
\usepackage{tablefootnote}
\usepackage[dvipsnames]{xcolor}
\usepackage{svg}
\newcommand{\cmmnt}[1]{}

\algnewcommand\algorithmicforeach{\textbf{for each}}
\algdef{S}[FOR]{ForEach}[1]{\algorithmicforeach\ #1\ \algorithmicdo}

\newcommand{\D}{\mathcal{D}}

\DeclareMathOperator*{\argmax}{arg\,max}
\DeclareMathOperator*{\EO}{\mathbb{E}}

\title{\LARGE \bf
A Model-Based Approach for Improving Reinforcement Learning Efficiency Leveraging Expert Observations
}

\author{Erhan Can Ozcan$^{1}$, Vittorio Giammarino$^{1}$, James Queeney$^{2}$, and Ioannis Ch. Paschalidis$^{3}$
\thanks{ECO, VG, and ICP were partially supported by the DOE under grants DE-AC02-05CH11231 and DE\-EE0009696, the NSF under grants CCF\-2200052, DMS\-1664644, and IIS\-1914792, and by the ONR under grants N00014\-19\-1\-2571 and N00014\-21\-1\-2844. JQ was exclusively supported by Mitsubishi Electric Research Laboratories (MERL).}
\thanks{A shorter version of this work has been submitted to the IEEE for possible publication. Copyright may be transferred without notice, after which this version may no longer be accessible.}
\thanks{$^{1}$Erhan Can Ozcan and Vittorio Giammarino are with Division of Systems Engineering, Boston University, Boston, MA 02215, USA
        {\tt\small \{cozcan,vgiammar\}@bu.edu}}%
\thanks{$^{2}$James Queeney is with Mitsubishi Electric Research Laboratories (MERL), Cambridge, MA 02139, USA
        {\tt\small queeney@merl.com}}%
\thanks{$^{3}$Ioannis Ch. Paschalidis is with the Department of Electrical and Computer Engineering, Division of Systems Engineering, and Department of Biomedical Engineering, Boston University, Boston, MA 02215, USA
        {\tt\small yannisp@bu.edu}}%
}

\begin{document}

\maketitle
\thispagestyle{empty}
\pagestyle{empty}

\begin{abstract}

\cmmnt{Reinforcement learning (RL) has proved its capability by performing extremely well on various control tasks. However, the biggest challenge for the deployment of RL algorithms in real-world problems is their high sample complexity. On the other hand, imitation learning (IL) can approximate the expert performance by leveraging expert demonstrations. Although approaches from both fields may have their own limitations when deployed individually, in some cases, it is possible to combine both disciplines to overcome these shortcomings, thereby increasing overall performance. Unlike existing literature,}
This paper investigates how to incorporate expert observations (without explicit information on expert actions) into a deep reinforcement learning setting to improve sample efficiency. First, we formulate an augmented policy loss combining a maximum entropy reinforcement learning objective with a behavioral cloning loss that leverages a forward dynamics model. Then, we propose an algorithm that automatically adjusts the weights of each component in the augmented loss function. Experiments on a variety of continuous control tasks demonstrate that the proposed algorithm outperforms various benchmarks by effectively utilizing available expert observations.
\end{abstract}

\section{INTRODUCTION}

{\em{Deep Reinforcement Learning}} (RL) has shown its potential to solve complex sequential decision making problems with its impressive success in learning to play video games \cite{mnih2013playing}, autonomous driving \cite{kiran2021deep}, and robotics applications \cite{ravichandar2020recent}. Among RL algorithms, model-free approaches are appealing  since they can serve as general-purpose tools for learning complex tasks \cite{lillicrap2015continuous,abdolmaleki2018maximum}. However, these approaches are generally not sample efficient, and often require a significant number of interactions with the environment during learning. Since this process can be costly, time-consuming, and unsafe, the use of these algorithms can be limited in real-world problems \cite{queeney2021generalized, queeney2023optimal}.\cmmnt{ data during learning, particularly in cases where the target task is complex \cite{torabi2018behavioral}. Since data collection can be both costly and time-consuming, the use of these algorithms can be limited in real-world problems \cite{queeney2021generalized}.}

The efficiency issue of RL algorithms can be mitigated by leveraging available expert data. Unfortunately, expert data contain limited information, possibly having only access to observations of the expert (e.g., states), but not to their private decision making (e.g., actions). This  mandates estimating expert actions or rewards before employing an RL algorithm to learn a policy \cite{liu2018imitation, giammarino2023opportunities}. Since the performance of these types of algorithms can be susceptible to estimation errors, leveraging expert observations remains as a challenging task \cite{prudencio2023survey,zheng2021imitation}.

In this study, we focus on improving the performance of an RL algorithm by leveraging expert observations. First, we define a policy loss function by combining a maximum entropy RL objective with a behavioral cloning loss to imitate the state transitions of the expert. This behavioral cloning loss leverages a forward dynamics model, which is learned in an online manner, to address the absence of expert actions. Next, to mitigate the detrimental effect that an inaccurate forward model can have on policy learning, we propose a practical algorithm that automatically adjusts the weight of the behavioral cloning loss during policy learning based on the accuracy of the model. Finally, by conducting experiments in the DeepMind Control Suite \cite{tassa2018deepmind}, we demonstrate that utilizing available expert data can significantly accelerate the training of RL algorithms.

\section{RELATED WORK}
One possible way to simplify policy learning in RL is to consider a set of expert data and utilize offline RL algorithms \cite{rajeswaran2017learning, gao2018reinforcement, nair2020awac,xu2021offline}. However, all these studies assume that the expert dataset is collected in a demonstration format, i.e., sequences of states, actions, and reward tuples, which is unrealistic in many scenarios. Therefore, it is important to consider alternative algorithms that can work with expert datasets containing limited information. 

{\em{Imitation Learning}} (IL) is another field that investigates how to incorporate expert data into policy learning in the absence of reward information. The goal in {\em{Inverse Reinforcement Learning}} (IRL), a popular sub-field of IL, is to retrieve a reward signal based on the expert data and utilize this information to learn a policy \cite{zheng2021imitation}. Various IRL approaches utilize the expert's state-action pairs \cite{finn2016guided, ho2016generative, baram2016model}. 
However, data showing both expert states and actions on a specific task can be either limited or unavailable, which hinders the deployment of these types of algorithms in real-world problems. On the contrary, expert observations can be easily accessible in today's world, thus attention has shifted towards studies that focus on learning a policy by using only expert states. In \cite{torabi2018generative}, Generative Adversarial Imitation Learning proposed in \cite{ho2016generative} has been extended to a state-only setting. In \cite{sun2019provably}, an algorithm that relies only on expert states has been developed to learn a non-stationary policy for discrete actions. In \cite{kidambi2021mobile}, a model-based framework, which promotes exploration in the face of uncertainty, is proposed to learn a policy solely from expert states. Finally, two recent works explore the possibility of adversarial imitation from expert's visual observations \cite{giammarino2023adversarial, liu2023visual}. However, all these approaches optimize an adversarial min-max problem during policy learning since they lack a reward signal, making optimization difficult \cite{zheng2021imitation,tucker2018inverse,brown2019extrapolating}.\cmmnt{Consequently, the use of these approaches can be limited in high-dimensional tasks \cite{zheng2021imitation,tucker2018inverse,brown2019extrapolating}.}

{\em{Behavioral Cloning}} (BC) is another sub-field of IL, where the goal is to learn a direct mapping from states to actions in a supervised manner. In this field, one group of work assumes that there exists an expert providing the optimal action for any given state, and propose BC approaches based on dataset aggregation \cite{ross2011reduction,kim2013maximum}. However, having this type of an expert can be either extremely costly or unrealistic in many cases. In \cite{zheng2021imitation, hanawal2019learning}, BC algorithms that can be trained on offline expert demonstrations are outlined, where the agent does not need to interact with the environment during policy learning, but requiring access to expert actions along with the states. Similar to IRL, studies that focus on learning a policy by using only expert states are of interest in BC. In \cite{edwards2019imitating}, a BC approach is introduced where an expert can be imitated by learning a latent forward dynamics model. Finally, in \cite{torabi2018behavioral}, a BC approach is proposed that utilizes only expert states. Unfortunately, pure BC approaches may suffer from compounding errors as a result of distribution shift \cite{ross2011reduction,laskey2016shiv}. Therefore, the performance of BC can be poor in domains where consecutive actions are required for success \cite{torabi2018generative}.

In order to address the shortcomings of both IL and RL approaches, combining these two disciplines is an appealing strategy when possible, i.e., a reward signal can be received from the environment after the interaction. In \cite{cheng2018fast}, a first-order oracle provides policy update information for any given state during the policy imitation stage, which is followed by traditional RL steps. In \cite{giammarino2022combining}, policy learning begins with IL on available expert demonstrations, and then the learned policy is refined via RL to achieve a human level performance. Finally, in \cite{goecks2019integrating}, an actor-critic algorithm is proposed by optimizing an RL objective combined with a behavioral cloning loss defined on expert demonstrations. This is similar to the augmented loss considered in this work, but the approach in \cite{goecks2019integrating} utilizes expert actions during policy training that we assume are not available.

State-only expert data can be incorporated into policy optimization by learning a world model, which reflects the transition dynamics\cmmnt{, and the learned model plays a key role in the success of the RL algorithm. Therefore, selecting the right model is of utmost importance \cite{chua2018deep}. While non-parametric models such as Gaussian processes are favourable in low-dimensional settings \cite{ko2007gaussian}, \cite{kumar2016optimal}, \cite{kamthe2018data}, the deployment of these models in real-world problems can be limited as they lack expressiveness \cite{kurutach2018model}. On the other hand,}. Neural networks are powerful function approximators that can model complex non-linear dynamics even in high-dimensional problems; thus, they can be good candidates for modeling the dynamics in real-world problems \cite{agrawal2016learning}, \cite{nagabandi2020deep}. Most of the existing model-based approaches utilize the learned model as a simulator to generate huge amounts of data, which is how they achieve sample efficiency during policy optimization \cite{rajeswaran2020game, janner2019trust}. One common drawback of this strategy is the challenge of determining the length of the counterfactual trajectories generated by the model since the model errors accumulate as the model horizon increases.\cmmnt{Therefore,  In our work, the learned predictive model facilitates the expert state-only demonstrations to improve policy learning.} Therefore, contrary to existing work, we utilize the learned predictive model solely to facilitate the use of expert observations in our work.

\section{PRELIMINARIES}
\label{sec_pre}

Consider an infinite horizon {\em{Markov Decision Process}} (MDP) characterized by the tuple $(\mathcal{S},\mathcal{A},p,r,\gamma,\rho_0)$. Following the standard notation, $\mathcal{S}$ and $\mathcal{A}$ denote the continuous state and action spaces, respectively, $p: \mathcal{S} \times \mathcal{A} \rightarrow P(\mathcal{S})$ is the transition model where $P(\mathcal{S})$ denotes the space of probability measures over $\mathcal{S}$, $r: \mathcal{S} \times \mathcal{A} \rightarrow \mathbb{R}$ is the reward function, $\gamma\ \in\ (0,1)$ is the discount rate, and $\rho_0$ is the initial state distribution.

The agent interacts with the environment by following a stochastic policy parameterized by $\theta$, $\pi_\theta: \mathcal{S} \rightarrow P(\mathcal{A})$, and the expected sum of discounted rewards $\eta(\pi)$ can be expressed as follows:

\[\eta(\pi)=\mathbb{E}_{(s_t,a_t)\sim \rho_{\pi}}\left[\sum_{t=0}^\infty\gamma^tr(s_t,a_t)\right],\]
where $(s_t,a_t)\sim\rho_{\pi}$ denotes a trajectory sampled according to  $s_0 \sim \rho_0$, $a_t \sim \pi_\theta(\, \cdot \mid s_t)$, and $s_{t+1} \sim p(\, \cdot \mid s_t,a_t)$.

\cmmnt{Then, the goal of the standard reinforcement learning is to find a policy $\pi^*$ maximizing the expected sum of discounted rewards over a policy class $\Pi$:
\[\pi^* = \argmax_{\pi \in \Pi}\ \eta(\pi).\]}
Then, the goal of standard RL is to find a policy maximizing the expected sum of discounted rewards. By augmenting the standard reinforcement learning objective with an entropy term, the maximum entropy objective can be obtained \cite{ziebart2010modeling}. Accordingly, the optimal policy $\pi^*$ of this problem maximizes the entropy visited at each state as well as the discounted reward:
\[\pi^* = \arg \max_{\pi} \ \mathbb{E}_{(s_t,a_t)\sim \rho_{\pi}}\left[\sum_{t=0}^\infty\gamma^tr(s_t,a_t) +\alpha H(\pi(\cdot\ | s_t))\right],\]
where $H$ represents entropy and $\alpha$ is the temperature parameter adjusting the relative weight of the entropy term with respect to the discounted reward. {\em{Soft Actor-Critic}} (SAC) \cite{haarnoja2018soft} is an effective off-policy algorithm within this maximum entropy RL framework, and it alternates between estimating a soft Q-value function $Q_{\psi}(s,a)$ using the soft Bellman backup operator and a policy improvement step to improve the current policy $\pi_\theta$. Using a replay buffer $\mathcal{D}$ that stores previously sampled states and actions, the soft Q-value parameters, $\psi$, can be obtained by optimizing the following soft Bellman residual error:
\begin{equation}
\begin{split}
&J_Q(\psi,\D) = \EO_{(s_t,a_t)\sim \D}\left[\frac{1}{2}(Q_{\psi}(s_t,a_t) - (r_t + \gamma V_{\bar{\psi}}(s_{t+1}) ))^2\right],\\
&\text{where}\\
&V_{\bar{\psi}}(s_{t+1})= \EO_{a_{t+1}\sim \pi_\theta}[Q_{\bar{\psi}}(s_{t+1},a_{t+1})-\alpha \log\pi_\theta(a_{t+1} | s_{t+1})], \label{soft_bellman}
\end{split}
\end{equation}
and $\bar{\psi}$ are delayed target parameters that are maintained via exponential smoothing with coefficient $\tau$. Then, SAC updates the policy towards the exponential of soft Q-value based on the following maximum entropy policy loss:
\begin{equation}
\label{policy_loss}
J_\pi(\theta,\D) = \mathbb{E}_{s_t\sim \D, a_t \sim \pi_\theta}\left[\alpha\log\pi_\theta(a_t | s_t) - Q_{\psi}(s_t,a_t)\right].
\end{equation}
The final ingredient in SAC is to devise a mechanism to automate the temperature selection for each task based on a target entropy value $\bar{H}$. Therefore, $\alpha$ is updated based on the gradients of the following function:
\begin{equation}
\label{temperature_update}
J(\alpha, \D) = \mathbb{E}_{s_t\sim \D, a_t \sim \pi_\theta} \left[-\alpha\log\pi_\theta(a_t | s_t) -\alpha\bar{{H}}  \right].
\end{equation}

The algorithm proposed in this paper builds on top of this maximum entropy RL framework. While we incorporate the same Q-value estimation step and temperature update rule into our algorithm, we modify the policy loss given in \eqref{policy_loss} to utilize expert observations during policy updates, thereby significantly accelerating training.


\section{MAXIMUM ENTROPY POLICY LEARNING WITH EXPERT OBSERVATIONS}

We now present our algorithm \emph{Soft Actor-Critic with Expert Observations} (SAC-EO), which learns a policy with maximum entropy RL while also minimizing deviations from the expert's state trajectory. The algorithm has two essential components: (1) We learn a forward dynamics model that allows us to utilize expert observations to improve policy learning without requiring expert actions, and (2) we propose an augmented policy objective that combines the maximum entropy RL objective with a behavioral cloning loss, leveraging expert observations.

\paragraph*{\textbf{Incorporation of Expert Data}} Although we only observe the expert states, it is still possible to utilize this information during policy optimization by learning a forward dynamics model. Learning a global model that can accurately predict the next state and the reward for any given state-action pair can be impractical, as it requires exploring all parts of the state space. Furthermore, learning a globally accurate model can be impossible when the system dynamics are complex \cite{rajeswaran2020game}. However, this is not necessary for the success of our algorithm since we employ our model solely to incorporate expert data into the policy improvement.

Our goal is to modify the policy $\pi_\theta$ by utilizing a learned model so that the policy can generate transitions similar to the expert. We model the transition dynamics using a probabilistic feed-forward neural network parameterized by $\phi$, whose outputs represent the mean vector $\mu$ and diagonal covariance matrix $\Sigma$ of a Gaussian distribution. Then, for a state-action pair $(s_t,a_t)$ at time $t$, the trained model $\widehat{m}_\phi$ estimates the state at time $t+1$: \[\widehat{s}_{t+1} \sim \widehat{m}_\phi(s_t,a_t) = \mathcal{N}\left(\mu_\phi(s_t,a_t),\Sigma_\phi(s_t,a_t)\right).\] Similar to model-based RL algorithms, we refine the learned model gradually as the policy improves to overcome the state {\em{distribution shift}} issue \cite{rajeswaran2020game,janner2019trust}; more details related to model learning can be found in Appendix \ref{SACED_training}\cmmnt{\cite{this_paper_longer}}.

Suppose we are provided a set of expert observations (states) $\mathcal{D}^e \equiv \{s^e_t\}_{t=0}^N$. Then, for any given expert state $s^e_t$, we can estimate the next state of our policy by utilizing the learned model: $\widehat{s}_{t+1} \sim \widehat{m}_\phi(s^e_t, \pi_\theta(s^e_t)).$ Note that $\widehat{s}_{t+1}$ is a function of the policy parameters $\theta$, and our goal is to imitate the state transitions of the expert. Therefore, for a given model, policy and expert dataset, we can define a term measuring the deviation from the state trajectory of the expert by using our model:
\begin{equation}
\label{mse}
    \text{MSE}(\phi,\theta,\mathcal{D}^e)=\frac{\sum_{t=0}^{N-1}\|\widehat{s}_{t+1}-s^e_{t+1}\|^2}{N},
\end{equation}
where $\|\cdot\|$ denotes the $\ell_2$ norm of a vector. Since we learn a parametric policy and model, we can calculate the gradient of the MSE term with respect to $\theta$, and utilize this information during policy optimization. Finally, note that our learned model is only used to calculate the MSE term in \eqref{mse}, unlike model-based RL methods that utilize the learned model as a simulator to generate additional data for training.

\paragraph*{\textbf{Augmented Policy Objective}} Similar to \cite{haarnoja2018soft}, our algorithm optimizes an RL objective, namely the maximum entropy RL objective given in \eqref{policy_loss}, based on data available in the replay buffer $\D$. In addition to this, our algorithm takes into account the expert data with the help of the learned model, and minimizes the deviations from the state trajectory of the expert defined in \eqref{mse}. The augmented function minimized by our algorithm at the policy improvement stage is as follows:
\begin{equation}
    \label{const_eps}J_\pi(\theta,\phi,\D,\mathcal{D}^e,\epsilon)=(1-\epsilon)J_\pi(\theta,\D) + \epsilon\text{MSE}(\phi,\theta,\mathcal{D}^e),
\end{equation}
where $\epsilon$ is the expert state matching parameter controlling the relative importance of the MSE term compared to the soft
policy improvement term. Next, we describe how to automatically adjust $\epsilon$ throughout training.

\subsection{Automatic Adjustment of the Expert State Matching Coefficient}

The ingredients above describe how we can incorporate expert observations into a maximum entropy RL algorithm. Nevertheless, the success of our algorithm heavily depends on the quality of the learned model, which is used to calculate the MSE term in \eqref{const_eps}. In the early stages of training, the learned model may not be able to accurately represent transition dynamics in states visited by the expert. \cmmnt{On the contrary, the state distribution induced by the policy shifts toward the expert's state distribution as the policy improves. Hence, the model represents the expert dynamics more accurately as training progresses, and information coming from the MSE term becomes more valuable.}In addition, it may be difficult to learn an accurate model in environments with complex dynamics. In these scenarios, we want to place more emphasis on the RL objective. 

To integrate this concept into our framework, it is important to devise a strategy that determines the importance of each term based on the accuracy of the model on the expert dataset, and automatically adjusts the value of the expert state matching parameter. However, evaluating the performance of the model on the expert trajectories requires knowledge of both expert actions and states. Since we only have access to expert states, evaluating the performance of the model is not trivial.

In model-based RL frameworks, it is common to learn a model ensemble and to define a discrepancy measure between models to assess their accuracy. For example, \cite{kidambi2020morel} utilizes a discrepancy measure to prevent policies from visiting states where transition dynamics are uncertain, while \cite{kidambi2021mobile} utilizes a discrepancy measure to incentivize the policy to explore unknown parts of the state space. In our study, we want to assess the reliability of our models on the states visited by the expert. Therefore, we design a rule based on this measure.

Suppose that we have a pair of dynamics models, $\widehat{m}_{\phi_1}$ and $\widehat{m}_{\phi_2}$, that are initialized with the parameters $\phi_1$ and $\phi_2$, respectively, but trained on the same data. We can define the maximum discrepancy $\delta^{max}$ between models on expert dataset $\D^e$ as follows:
\begin{equation}
    \label{delta_max}\delta^{max}(\D^e)=\max_{s \in \D^e}\ \|\widehat{m}_{\phi_1}(s, \pi_\theta(s)) - \widehat{m}_{\phi_2}(s, \pi_\theta(s))\|.
\end{equation}
Using this discrepancy measure, we can define an adaptive expert state matching coefficient by
\begin{equation}
\epsilon_k = \frac{1}{1+\beta \delta^{max}(\D^e)},
\end{equation}
where $\beta \geq 0$ is a tunable parameter to scale the discrepancy. Then, the policy loss with an adaptive epsilon $\epsilon_k$ can be expressed as follows:
\begin{equation}
\begin{split}
    \label{adpt_eps}J_\pi(\theta,&\phi,\D,\mathcal{D}^e,\epsilon_k)=\\&(1-\epsilon_k)J_\pi(\theta,\D)+ 
                                                    \epsilon_k\frac{\sum_{i=1}^2\text{MSE}(\phi_i,\theta,\mathcal{D}^e)}{2}.
\end{split}
\end{equation}
Finally, we assume that our algorithm is trained by collecting $K$ trajectories, each of length $E$, and we update the models and adaptive state matching coefficient at the end of each trajectory. The pseudocode showing the steps of SAC-EO is given in Algorithm \ref{alg:one}.


\begin{algorithm}[ht]
\caption{SAC-EO}
\label{alg:one}
\hspace*{\algorithmicindent}\textbf{Input:} Scale parameter $\beta \geq 0$, Expert observations $\D^e$. 
\begin{algorithmic}[1]
\State Initialize policy $\pi_\theta$, predictive models $\widehat{m}_{\phi_1}$ and $\widehat{m}_{\phi_2}$, critics $Q_{\psi_1}$ and $Q_{\psi_2}$, target critics $Q_{\overline{\psi}_1}$ and $Q_{\overline{\psi}_2}$, replay buffer $\mathcal{D}$, model training buffer $\mathcal{D}^{m}$.
\For{\texttt{$k=1\ ...\ K$}}
        \State Train models $\widehat{m}_{\phi_1}$ and $\widehat{m}_{\phi_2}$ on $\D^{m}$ via maximum likelihood.
        \State $\delta^{max}=\max_{s \in \mathcal{D}^e} \|\widehat{m}_{\phi_1}(s, \pi_\theta(s)) - \widehat{m}_{\phi_2}(s, \pi_\theta(s))\|$.
        \State Set $\epsilon_k=\frac{1}{1+\beta \delta^{max}}$.
        \State Sample a mini-batch of expert states from $\D^e$, and store them in $\D^e_{b}$.
        \For{\texttt{$E$ steps}}
            \State $a_t \sim \pi_\theta(\cdot|s_t)$
            \State $s_{t+1}, r_t \sim$ EnvironmentStep$(a_t)$
            \State $\D \leftarrow \D \cup (s_t,a_t,r_t,s_{t+1})$
            \State $\D^m \leftarrow \D^m \cup (s_t,a_t,s_{t+1})$
            \State Sample a mini-batch of environment data from $\D$, and store them in $\D_{b}$.
            \State Split $\mathcal{D}^e_{b}$ into two parts randomly: $\mathcal{D}^e_{b,1}$ and $\mathcal{D}^e_{b, 2}$.
            \State $\psi_i \leftarrow \psi_i - \lambda_Q\nabla_{\psi_i}J_Q(\psi_i,\D_{b})\ \text{for } i \in \{1,2\}$
            \State $L_\theta=(1-\epsilon_k)J_\pi(\theta,\D_{b}) + \epsilon_k\frac{\sum_{i=1}^2\text{MSE}(\phi_i,\theta,\mathcal{D}^e_{b, i})}{2}$
            \State $\theta \leftarrow \theta - \lambda_\pi\nabla_\theta L_\theta$
            \State $\alpha \leftarrow \alpha - \lambda\nabla_\alpha J(\alpha, \D_b)$
            \State $\overline{\psi_i} \leftarrow \tau\psi_i + (1-\tau)\overline{\psi}_i\ \text{for } i \in \{1,2\}$


        \EndFor
\EndFor
\State \Return $\pi_\theta$.
\end{algorithmic}
\end{algorithm}

Note that setting $\beta=0$ in Algorithm \ref{alg:one} results in a modified version of the algorithm {\em{Behavioral Cloning From Observation}} (BCO) proposed in \cite{torabi2018behavioral}, and the algorithm converges to classic SAC as $\beta \rightarrow \infty$.

\section{Experiments}

We analyze the use of expert observations within a maximum entropy RL framework over 6 tasks available in the DeepMind Control Suite \cite{tassa2018deepmind}. The tasks we consider include {\em{cheetah-run}}, {\em{hopper-hop}}, {\em{hopper-stand}}, {\em{walker-run}}, {\em{walker-walk}}, and {\em{walker-stand}}, which all have fixed-length horizons of 1000 steps with rewards in the unit interval, i.e., $r(s,a)\ \in\ [0,1]$. We obtain expert observations by training a different model-free RL algorithm {\em{Maximum a Posteriori Policy Optimization}} (MPO) \cite{abdolmaleki2018maximum} for 5 million steps. The full details of SAC-EO algorithm, including network architectures, model training, and all hyperparameters are summarized in \cmmnt{\cite{this_paper_longer}}Appendix \ref{SACED_training}. Finally, the source code is available on GitHub to promote reproducibility\footnote{\label{note_one}\url{https://github.com/noc-lab/sac-expert}.}.

\begin{figure*}[!h]
\centering
\includegraphics[width=7.5in]{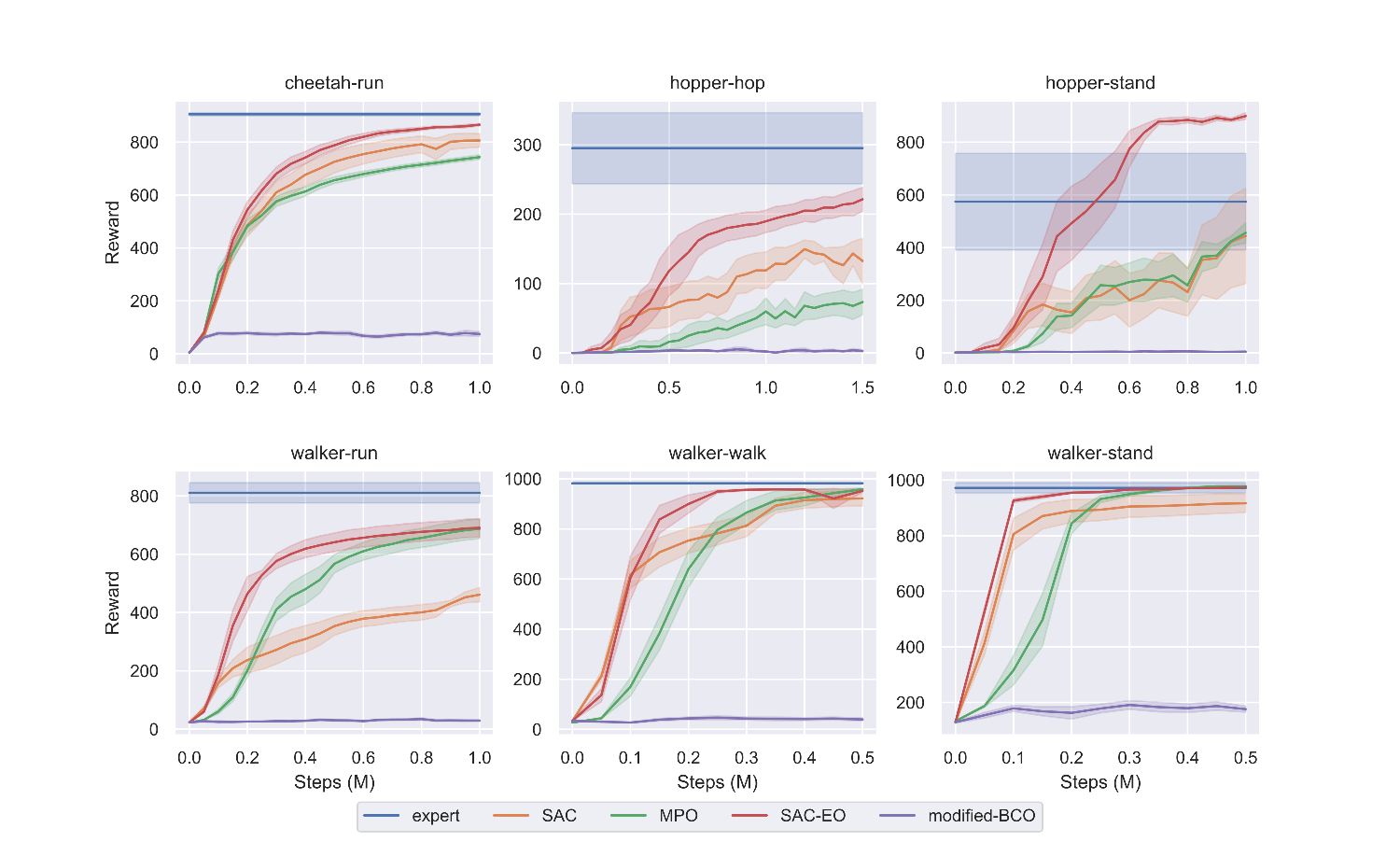}
\caption{Comparison of algorithms across tasks. Horizontal blue line represents the expert performance. SAC-EO and modified-BCO are supplied four expert trajectories during training. The scale parameter of SAC-EO is the same across all tasks. Shading denotes one standard error across policies.}
\label{fig:one}
\end{figure*}

\subsection{Performance Comparison on DeepMind Control Suite}

Although our strategy to incorporate expert data into policy improvement can be generalized into any model-free RL algorithm, we build our algorithm on top of SAC due to its flexibility to learn complex tasks. Therefore, the first benchmark we consider in our experiments is SAC. Another benchmark we consider in our experiments is MPO since expert data required by our algorithm are generated by employing an MPO agent. We consider a problem setting where only expert states are available during training, thus offline RL algorithms cannot be used as baselines. However, we include a modified version of BCO, obtained by setting $\beta=0$ in Algorithm \ref{alg:one}, among the benchmarks for a fair comparison. { The reason to use modified-BCO is to ensure a one to one correspondence between methods, and one fundamental difference between modified-BCO and BCO is that modified-BCO utilizes a forward dynamics model instead of an inverse model during policy optimization.} Figure \ref{fig:one} shows the average of total reward obtained by our algorithm across 5 seeds on different tasks, compared to SAC, MPO, and modified-BCO.

According to Figure \ref{fig:one}, SAC-EO matches the performance of the expert, which was trained for five millions steps, in less than one million steps for 4 out of 6 tasks. Furthermore, utilizing expert observations during policy updates accelerates the training of SAC in all tasks. We further analyze this point in Table \ref{table:two}, which reports the number of steps each algorithm needs to reach a percentage of the expert performance. Lastly, the poor performance of modified-BCO in all tasks reveals that the behavioral cloning loss defined on expert observations alone is insufficient to learn a good policy, since it is not possible to learn a model that reflects the transition dynamics on the expert states unless the agent explores this part of the state space. This highlights the importance of utilizing this type of a loss within an RL framework.

\begin{table}[htp]
\centering
\caption{Steps to reach X percentage of the expert performance\tablefootnote{In Table \ref{table:two}, NA means that the specified percentage of the expert performance cannot be achieved by the corresponding algorithm in one million steps.}}\label{table:two}
\begin{tabular}{ l *{6}{c} }
\toprule
\hspace*{2cm} &
\multicolumn{3}{c}{75\% of the Expert} &
\multicolumn{3}{c}{95\% of the Expert} \\
\cmidrule(lr){2-4} \cmidrule(lr){5-7}
&
\makebox[2em]{MPO} &
\makebox[2em]{SAC} &
\makebox[2em]{SAC-EO} &
\makebox[2em]{MPO} &
\makebox[2em]{SAC} &
\makebox[2em]{SAC-EO} \\
\midrule
cheetah-run & 577K & 368K & \textbf{265K} & NA & NA & \textbf{770K} \\
hopper-hop & NA & NA & NA & NA & NA & NA \\
hopper-stand & 508K & 250K & \textbf{227K} & 799K & 627K & \textbf{287K} \\
walker-run & 571K & NA & \textbf{330K} & NA & NA & NA \\
walker-walk & 190K & 139K & \textbf{83K} & 376K & 357K & \textbf{185K} \\
walker-stand & 160K & 47K & \textbf{29K} & 187K & 268K & \textbf{48K} \\
\bottomrule
\end{tabular}
\end{table}

\subsection {The Benefit of Adaptive Epsilon}

{Next, we show the benefit of utilizing an adaptive $\epsilon_k$, as we describe, over a fixed $\epsilon_k$ in our algorithm. Therefore, we train our algorithm by setting $\epsilon_k$ to $0.1,\ 0.01,$ and $0.001$, respectively, and compare them with our adaptive algorithm. The average performance of each version after one million interactions with the environment is summarized in Table \ref{table:adaptive}.}

\begin{table}[!h]
\centering
\caption{The benefit of utilizing adaptive algorithm}\label{table:adaptive}
\scalebox{.95}{
\begin{tabular}{lcccc}
\hline
\textbf{\begin{tabular}[c]{@{}c@{}}Fixed vs Adaptive\end{tabular}} & \textbf{\begin{tabular}[c]{@{}c@{}}$\epsilon_k=0.1$\end{tabular}} & \textbf{\begin{tabular}[c]{@{}c@{}}$\epsilon_k=0.01$\end{tabular}} & \textbf{\begin{tabular}[c]{@{}c@{}}$\epsilon_k=0.001$\end{tabular}} & \textbf{\begin{tabular}[c]{@{}c@{}} $\normalfont{\text{Adaptive}}$ \end{tabular}} \\ \hline
   cheetah-run & 614$\pm$93                  & 808$\pm$33 & 843$\pm$34  & \textbf{867$\pm$11} \\
   hopper-hop & 38$\pm$77                  & 86$\pm$96 & \textbf{196$\pm$32}  & 191$\pm$42  \\
   hopper-stand & 910$\pm$24                  & 564$\pm$435 & 602$\pm$419  & \textbf{919$\pm$25} \\
   walker-run & 23$\pm$2                  & 253$\pm$169 & 623$\pm$127  & \textbf{695$\pm$62} \\
   walker-stand & 963$\pm$30                  & 981$\pm$9 & 982$\pm$9  & \textbf{985$\pm$10}\\
   walker-walk & 23$\pm$1                  & 948$\pm$28 & 955$\pm$41  & \textbf{967$\pm$11}  \\ \hline
\end{tabular}}
\end{table}

{According to Table \ref{table:adaptive}, it is promising to utilize our adaptive strategy, as the adaptive version performs better compared to using a fixed $\epsilon_k$ in most tasks. }

\section{CONCLUSION}

In this paper, we demonstrated how to increase the sample efficiency of an RL algorithm by utilizing expert observations without access to expert actions. We introduced the algorithm SAC-EO, which performs policy updates based on a loss function combining a maximum entropy RL objective with a behavioral cloning loss that utilizes a learned forward dynamics model. Based on our experiments, SAC-EO significantly accelerates training compared to standard model-free RL algorithms. Although our algorithm is built upon SAC, the strategy we present to incorporate expert observations into policy improvement is a general framework that can work with many RL algorithms. Therefore, it can be interesting to explore the effect of employing our idea on different RL algorithms in future work. Furthermore, since it does not need explicit action information from experts, the proposed algorithm may have the potential to take advantage of large amounts of video available online during policy improvement. Hence, another interesting future direction can be to extend our work by incorporating visual observations.

\bibliographystyle{IEEEtran} 
\bibliography{IEEEabrv,IEEEexample}

\newpage
\appendix

\subsection{Implementation Details and Hyperparameters}
\label{SACED_training}

In our experiments, the expert in each environment is trained on 5 million steps of environment interaction by using the algorithm MPO, and the trained experts are available in our public repository\footref{note_one}. Both policy and critics are represented by neural networks, which have 2 hidden layers of 256 units and RELU activations. After the first layer, a layer normalization followed by a tanh activation is applied. We consider a squashed multivariate Gaussian distribution with diagonal covariance as our policy class to bound actions into a finite interval. Therefore, given a state, the output of the policy network is the mean and the diagonal covariance of the action distribution. Finally, target versions of the critic networks are obtained as an exponential moving average of the weights with $\tau = 5\textnormal{e}{-3}$.


\begin{table}[H]
\caption{Network architectures and hyperparameters used to train SAC-EO}
\label{tab:saced}
\begin{center}
\begin{tabular}{lcc}
\toprule
\\
General &     \\
\cmidrule(lr){1-1}
Updates per environment step   && 1 \\ 
Discount rate ($\gamma$)        && 0.99 \\ 
Target network exponential moving average ($\tau$)        && $5\textnormal{e}{-3}$ \\  
Scale parameter ($\beta$) && 50, 100, 200 \\
Replay buffer size ($\D$) && $1\textnormal{e}{6}$ \\
Model buffer size ($\D^m$) && $1\textnormal{e}{5}$ \\
Number of available expert observations ($N$) && $4\textnormal{e}{3}$ \\
Number of trajectories ($K$) && $1\textnormal{e}{3}$ \\
Length of each trajectory ($E$) && $1\textnormal{e}{3}$\\[1.0em]

Policy &     \\
\cmidrule(lr){1-1}
Layer sizes       && 256, 256 \\ 
Layer activations && RELU \\ 
Layer norm + tanh on first layer && Yes \\ 
Initial standard deviation && 0.3 \\ 
Learning rate && $1\textnormal{e}{-4}$ \\ 
Entropy target ($\bar{H}$) && $-\text{dim}(\mathcal{A})$ \\
Initial temperature ($\alpha$) && $1\textnormal{e}{-1}$ \\
Mini-batch size to sample from $\D$ && 1024 \\
Mini-batch size to sample from $\D^e$ && 256 \\[1.0em]

Critics &     \\
\cmidrule(lr){1-1}
Layer sizes && 256, 256 \\ 
Layer activations && RELU \\ 
Layer norm + tanh on first layer && Yes \\ 
Learning rate && $3\textnormal{e}{-4}$ \\[1.0em] 

Models &     \\
\cmidrule(lr){1-1}
Layer sizes && 256, 256 \\ 
Layer activations && RELU \\ 
Layer norm + tanh on first layer && Yes \\ 
Learning rate && $1\textnormal{e}{-3}$ \\
Number of models && 2 \\
Mini-batch size to sample from $\D^m$ && 256 \\
Number of epochs && 10 \\[1.0em] 

\bottomrule
\end{tabular}
\end{center}
\end{table}

SAC-EO learns two Gaussian forward dynamics models to be able to utilize expert observations. Since expert actions are unknown, expert data cannot be utilized to train these models. Therefore, model training relies on the data collected from the environment by using the current policy. While the environment data is stored in the replay buffer $\D$, we utilize the buffer $\D^m$ to perform model updates. The same environment data is added to both $\D$ and $\D^m$, but the size of $\D^m$ is relatively smaller compared to $\D$. As a result of this, as the policy improves, the distribution of $\D^m$ changes more quickly, focusing more on the recently collected data to reflect expert dynamics better.


Lastly, SAC-EO adjusts the weight of the behavioral cloning loss during the policy improvement stage in an adaptive way. The adaptive $\epsilon_k$ has a scale parameter, $\beta$, that determines the contribution of each term in the aggregated policy loss function. During our experiments, we tune this parameter from a small set of values, which are 50, 100, and 200. Table~\ref{tab:saced} summarizes all important hyperparameter values used in our SAC-EO algorithm.

\end{document}